\documentclass[letterpaper, 10 pt, conference]{ieeeconf}  % Comment this line 

\IEEEoverridecommandlockouts                
\usepackage{amsmath}
\usepackage[ruled,vlined]{algorithm2e}
\usepackage{autolabtools}
\usepackage{tabu}
\usepackage{booktabs}
\usepackage{etoolbox}
\usepackage{graphics}
\usepackage{graphicx}
\usepackage{xcolor}
\usepackage{amsfonts}
\usepackage{hyperref}
\usepackage[font=small]{caption}
\usepackage{subcaption}
\usepackage{float}
\usepackage{adjustbox}

\usepackage{booktabs}
\usepackage{placeins}
\usepackage{epsfig} % for postscript graphics files
\usepackage{amssymb}  % assumes amsmath package installed
\usepackage{tikz}
\usepackage{standalone}
\usepackage{bondgraphs}
\usepackage{url}
\usepackage{blox}
\usepackage{textcomp}
\usepackage{gensymb}
\usepackage{booktabs}
\usepackage{tabularx}
\usepackage[keeplastbox]{flushend}
\usepackage{balance}
\usepackage{hyperref}
\usepackage{multirow}
\usepackage{bbm}
\let\labelindent\relax %needed for IEEE conflict with enumitem
\usepackage{enumitem} % for \begin{description}[align=left]

\DeclareUnicodeCharacter{0301}{``}

\usepackage[backend=biber,
            hyperref=true,
            url=false,
            isbn=false,
            doi=false,
            backref=false,
            style=ieee,
            natbib=true,%compatibility aliases
            mincitenames=1,
            maxcitenames=1,
            citestyle=numeric-comp,
            sorting=nyt,%none
            block=none]{biblatex}

\title{\LARGE \bf
Mechanical Search on Shelves using Lateral Access X-RAY
}

\author{Huang Huang*$^1$, Marcus Dominguez-Kuhne*$^{1,2}$, Jeffrey Ichnowski$^1$, Vishal Satish$^1$, Michael Danielczuk$^1$, \\
Kate Sanders$^1$, Andrew Lee$^1$,  Anelia Angelova$^3$, Vincent Vanhoucke$^3$, Ken Goldberg$^1$% <-this % stops a space

\thanks{*These authors contributed equally.
$^{1}$The AUTOLab at University of California, Berkeley.
$^{2}$The California Institute of Technology.
$^{3}$Robotics at Google.}
}

\begin{document}

\maketitle
\begin{abstract}
%Almost all objects are stored on shelves. <-- hard to justify, while good motivation, it is not the point of the paper -JI
Efficiently finding an occluded object with lateral access arises in many contexts such as warehouses, retail, healthcare, shipping, and homes.
We introduce LAX-RAY (Lateral Access maXimal Reduction of occupancY support Area), a system to automate the mechanical search for occluded objects on shelves. For such lateral access environments, LAX-RAY couples a perception pipeline predicting a target object occupancy support distribution with a mechanical search policy that sequentially selects occluding objects to push to the side to reveal the target as efficiently as possible.  
%We introduce LAX-RAY (maXimal Reduction of occupancY support Area in Lateral Access), a system to automate the mechanical search for occluded objects on shelves and other lateral access environments. 
% using LAX-RAY, an extension of the X-RAY system % \cite{danielczuk2020x}  -- don't put citations in abstracts, abstracts should stand alone. -JI
% Also, we shouldn't assume readers are familiar with X-RAY, we can reference it in the abstract, but only if we briefly describe what it does.
%to lateral access environments. 
% Based on the tables and the experiments, it seems like efficiency is just least number of steps? -MarcusDK 
Within the context of extruded polygonal objects and a stationary target with a known aspect ratio, we explore three lateral access search policies: Uniform, Distribution Area Reduction (DAR) and Distribution Entropy Reduction over n Steps (DER-n) utilizing the support distribution and prior information. We evaluate these policies using the First-Order Shelf Simulator (FOSS) in which we simulate 800 random shelf environments of varying difficulty, and in a physical shelf environment with a Fetch robot and an embedded PrimeSense RGBD Camera. 
Average simulation results of 87.3$\%$ success rate demonstrate better performance of DER-2. Physical results show a success rate of at least 80$\%$ for DAR and DER-n, suggesting that LAX-RAY can efficiently reveal the target object in reality. Both results show significantly better performance of DAR and DER-n compared to the uniform policy with uniform probability distribution assumption in non-trivial cases, suggesting the importance of distribution prediction. Code, videos, and supplementary material can be found at \url{https://sites.google.com/berkeley.edu/lax-ray}. 

% TODO: REMOVE ACRONYMS FROM THE ABSTRACT THAT ARE NO NEEDED.  KEEP "LAX-RAY" SINCE PEOPLE WILL SEARCH FOR IT AND ITS A CONTRIBUTION.  REMOVE DAR, DER, DER-MT, and FOSS, unless it makes a subsequent reference in the abstract easier (e.g., the physical experiments comparing DAR and DER).  I also worry that these names and abbreviations are not sufficient to convey the paper in an abstract.  -JI
\end{abstract}
\section{Introduction} \label{sec:introduction}
While researchers have explored the problem of mechanical search in unstructured clutter (in which objects have significant freedom in both position and orientation) \cite{danielczuk2020x,danielczuk2019mechanical}, mechanical search in structured, lateral access environments such as shelves, cabinets, and closets is a less studied area despite its wide applicability. For instance, a service robot at a pharmacy or hospital may need to find supplies from a cabinet, an industrial robot may need to find kitting tools from shelves in warehouses, or a service robot in a retail store may need to search shelves for requested products from customers. Searching for an object in a lateral access environment poses new challenges not faced in unstructured environments, namely, limited action spaces, complex motion planning requirements, and a potentially limited perception view.

Research on grasping objects in structured clutter, including in lateral access environments is also present in the literature ~\cite{murali20206, dogar2012physics, saxena2010vision}. However, few papers focus on the problem of searching for occluded objects in lateral access environments as opposed to grasping visible ones. A natural approach for searching for an object in lateral access environments is to iteratively look behind objects that could be occluding the target object. This can be improved through using an occupancy distribution identifying where an occluded target object could be hidden \cite{danielczuk2020x} and a policy estimating which objects should be pushed out of the way or removed given prior information about explored zones could greatly speed up the searching process. %In this paper, we propose Lateral-Access X-RAY (LAX-RAY) to automate search for occluded objects in lateral-access environments.

\begin{figure}
\vspace{4pt}
    \centering
    \includegraphics[width=.48\textwidth]{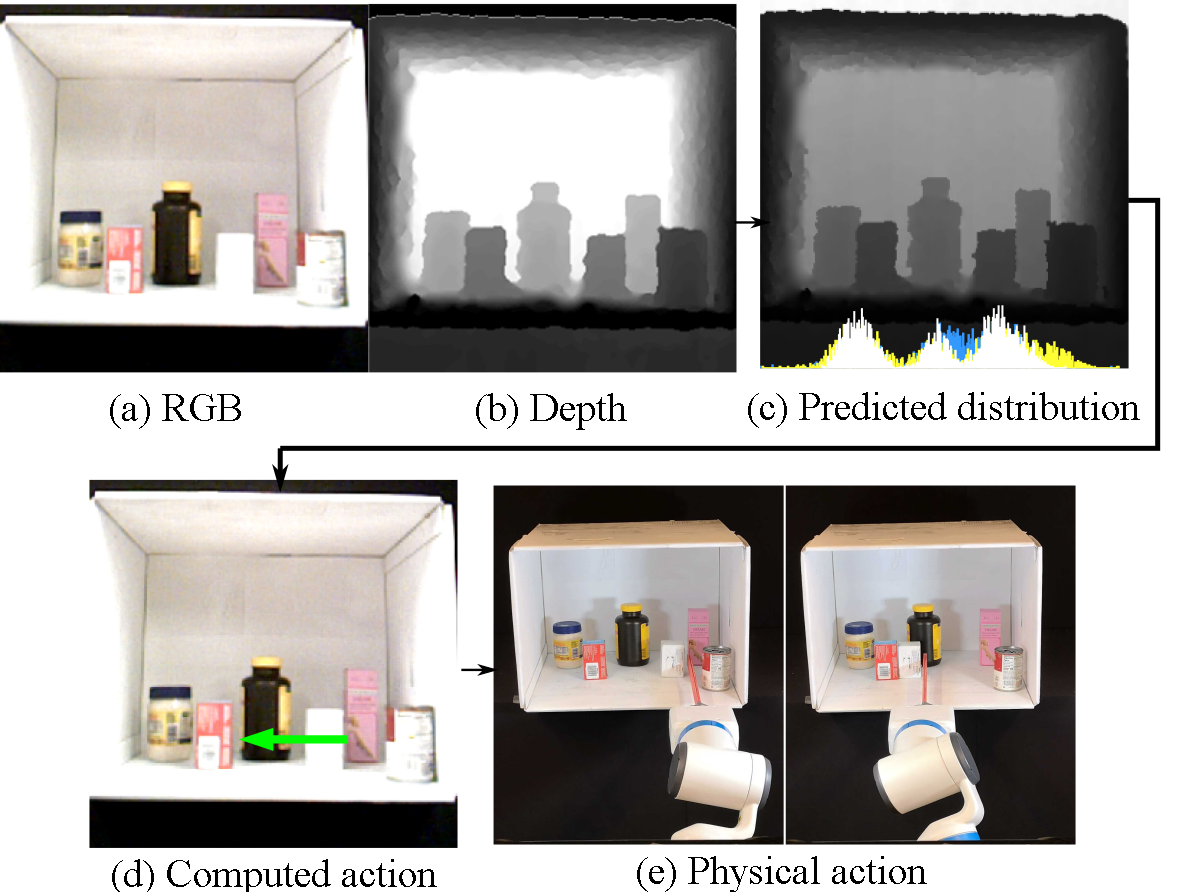}
    \caption{\textbf{Lateral-access mechanical search with LAX-RAY.}  %Diagram at time $t$ of Mechanical Search using LAX-RAY. 
    The search starts with an RGB Image (a) and Depth Image (b) of the environment. LAX-RAY perception predicts a distribution (c) of where the search target is hidden: at the current step (yellow), the previous time step (blue), and the minimum of the two (white). The policy computes a push action (d) indicated by the green arrow, and (e) the robot executes it.} %with the inputs of segmentation mask and probability distribution. In this diagram, line 1 is taking the minimum of current and history probability distribution. Line 2 is a pre-trained Fully Connected Network giving probability prediction based on the depth image and target object segmentation mask. Line 3 is a depth discontinuity based algorithm to give segmentation mask using the depth information and line 4 is a policy to give a pushing action based on the inputs. }
    \label{fig:diagram}
    \vspace{-12pt}
\end{figure}

This paper focuses on finding an occluded target object in a lateral access environment and considers a lateral-access variant of the \emph{Mechanical Search} problem~\cite{danielczuk2019mechanical}. In \cite{danielczuk2019mechanical}, the target object is in a bin and occluded from a top-down view by other objects. The lateral access environment introduces additional constraints including stable object poses due to gravity and limited operating space, where pushing operations can be more efficient.
% than grasping since removing is not required.
% and using pushes instead of grasps because object do not have to be removed in our setup. 
%
% Because lateral access changes the point of view and direction of gravity, the process for hypothesizing occluding objects, the search actions, and search policy must change. 
In this work, we extend X-RAY~\cite{danielczuk2020x}, a policy previously designed for bins with vertical access, to the lateral-access environment where all objects are resting in stable poses on a level shelf. We 
% push occluding objects away to avoid removing objects from the environment and 
work within a shelf as opposed to the infinite planar workspace X-RAY assumes. We adapt the occupancy distribution prediction method in \cite{danielczuk2020x} to the lateral access environment and additionally utilize prior information. Instead of grasping to remove objects on an infinite plane, we consider stable translation pushing actions in a tightly constrained shelf.

\begin{figure}[t]
\vspace{4pt}
    \centering
    \includegraphics[width=\columnwidth]{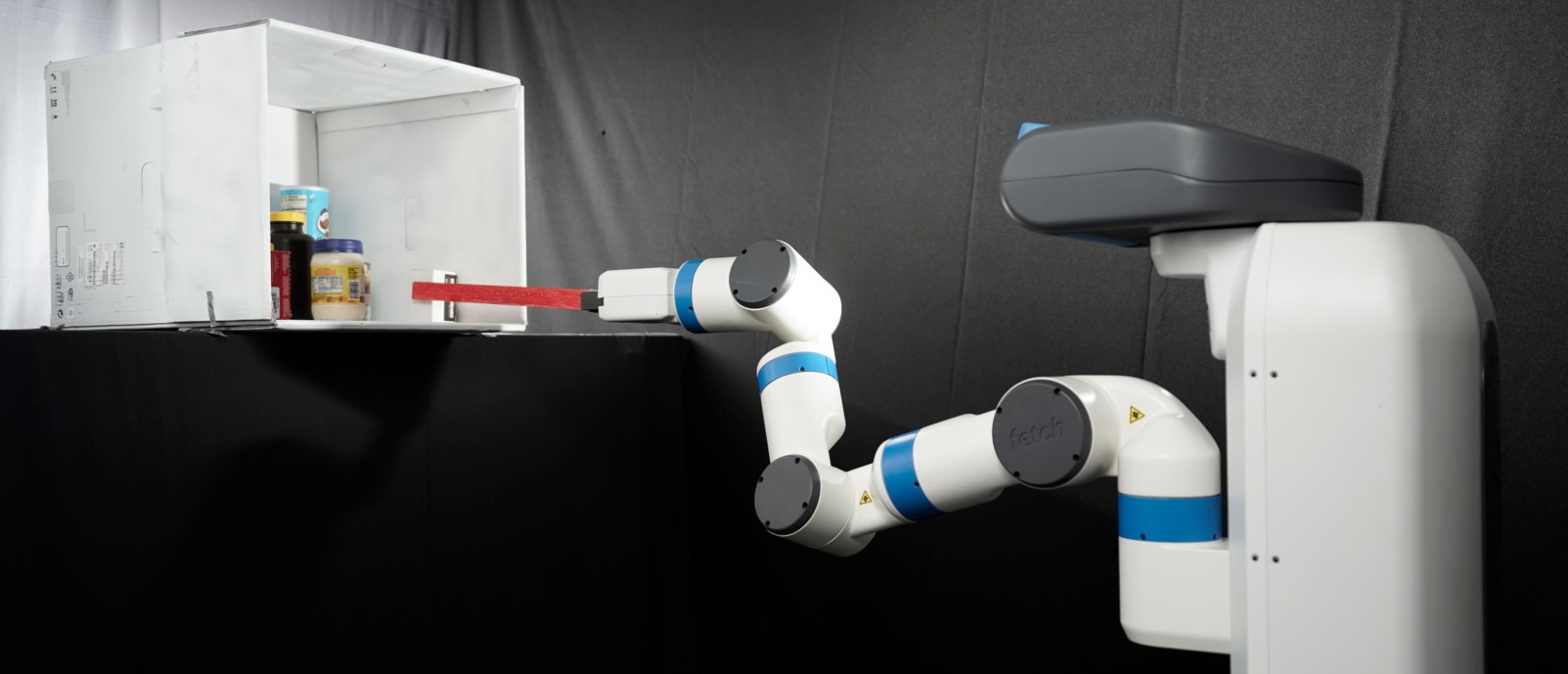}
    \caption{The physical experiment with a Fetch robot and a shelf environment with random objects. The robot pushes an object a distance and direction calculated by LAX-RAY taking inputs of a RGBD image from the embedded PrimeSense camera .}
    \label{fig:phy_setup}
    \vspace{-12pt}
\end{figure}

% While the constraints such as a restriction on viable poses and not removing objects impose new challenges, the lateral access environment is arguably more applicable in most real-world scenarios in which mechanical search would be useful.
This paper proposes LAX-RAY, a lateral access mechanical search system. Fig.~\ref{fig:diagram} shows a diagram of LAX-RAY. Fig. \ref{fig:phy_setup} shows an example of a push planned by LAX-RAY. The contributions of this study are as follows:
\begin{enumerate}
    \item An extension of the X-RAY deep-geometric inference system \cite{danielczuk2020x} to lateral-access environments predicting occupancy distribution.
    \item Three lateral access mechanical search policies: Uniform, DAR and DER-n ($\text{n}\in\{1,2,3\}$) that compute actions to reveal occluded target objects stored on shelves.
    \item The First Order Shelf Simulator (FOSS), a fast, open access, lightweight framework for generating initial shelf configurations and rapidly rolling out lateral search policies on them. 
    
    % \textcolor{red}{(include number of generated configs and rollouts)}. 
    \item Experiments in simulation and on a physical robot validating the policies. Results from 800 simulated and 5 physical trials suggesting that DAR and DER-2 have better performance with less and more objects respectively.
\end{enumerate}

\section{Related Work} \label{sec:relatedwork}

% {\color{red} [Can someone figure out how to make citet not allow breaks between "et al." and the number.  I find it super jarring to see lines begin with "[7]".  -JI]}

While we most notably take influence from work done in interactive perception and mechanical search \cite{zeng2020transporter, bohg2017interactive, van2014probabilistic, gupta2012using, danielczuk2019mechanical, kurenkov2020visuomotor}, work related to this problem spans back as far as the 1940s, such as Bayesian search: the problem of searching for one or more objects located in one of $m$ locations \cite{koopman1946search}.
% , was formulated during World War II .
\citet{assaf1985optimal} illustrate that, in many cases, this problem has an optimal greedy solution. Their work is expanded on by \citet{kress2008optimal}, \citet{lavis2008dynamic}, and \citet{wen2013sequential}, who consider cases of false-positive target detection, moving targets, and the production of sequences of optimal actions, respectively.

In robotics, \citet{lavalle1997motion} consider the problem of moving a target object through clutter without obscuring it behind other objects and propose online algorithms that attempt to maximize object visibility in future time steps. \citet{bitton2008hydra} illustrate methods for locating targets with multiple human and automated agents, and \citet{novkovic2020object} introduce a reinforcement learning-based system for exploring unknown environments and searching for target objects. \citet{yang2020deep} approach the search problem by using a target-centric motion critic that is trained on observations and metadata mapped to rewards, and \citet{murali20206} use a collision checking model to grasp objects in the way until the target object is uncovered. In this study, we avoid potential collisions by estimating the depth difference between objects.
% While we avoid grasping objects in this work due to cost, the principle of taking actions based on potential collisions to unocclude an object remains the same.

This task of searching for hidden objects is formalized and explored in the context of bin picking by \citet{danielczuk2019mechanical} and \citet{kurenkov2020visuomotor}. In their papers on mechanical search they consider the problem of uncovering, identifying, and extracting a target object from clutter. \citet{danielczuk2019mechanical} explore a set of policies that select pushing actions either randomly or by prioritizing objects to push based on factors such as size. These policies were improved by the X-RAY system \cite{danielczuk2020x}---a neural network approach trained on thousands of simulated observations and target object occupancy distributions that predicts which areas of a scene are most likely to occlude the target object.

% Object pose distribution modeling is also explored by \citet{prokudin2018deep}, \citet{kohl2018probabilistic}, \citet{price2019inferring}, and \citet{yang2019inferring}, who explore uncertainty quantification, generative segmentation models, completing shapes of occluded objects via neural network, and pixel-level depth distributions, respectively.

% To generate possible actions for object manipulation in the lateral access (shelf) environment, segmentation can determine objects in the scene. Many papers consider the applications and modifications to the problem of segmenting objects given an observational image. \cite{pinheiro2015learning, he2017mask, li2017fully, arnab2017pixelwise, wada2020instance, danielczuk2018segmenting} However, these papers do not consider the constraints on pose and structure of clutter that are considered in this paper.

As we specifically consider the task of extracting a target object from a lateral environment, pushing is a critical action for moving objects out of the way. Significant developments have recently been made in the study of singulating cluttered objects on a plane \cite{kopicki2009prediction, dogar2011framework, zito2012two, hermans2012guided, chang2012interactive, danielczuk2018linear, dong2019automating}. Notably, \citet{hermans2012guided} propose a pushing policy that includes a history of past actions used to determine when termination should occur. \citet{zeng2018learning} consider the problem of grasping in clutter by implementing robotic pushing and grasping actions and develop synergies between them through deep reinforcement learning. \citet{yang2020robotic} explore grasping in clutter through a pushing and grasping policy that makes use of a deep Q-learning network. This removes the need to generate training datasets as the training process for the policy is self-supervised. \citet{eitel2020learning} select push actions through the use of a neural network trained on physical data collected from robots interacting with cluttered scenes. The majority of the environments explored in these studies have relaxed constraints regarding object configurations and possible pushing actions compared to environments such as shelves, where a limited number of pushing actions are viable.

\section{Problem Statement}
\label{sec:problemstatement}
We consider an instance of the mechanical search problem in a lateral access environment (e.g., shelf).  In this problem, an agent searches for the target object that is in a stable pose on a shelf, occluded behind other objects on the shelf.  The target object has a known geometry, and the searching agent has RGBD camera with a fixed side-view of the shelf.  The agent can push objects on the shelf to the left and right to search for the object, but cannot remove objects.

We have the following assumptions: 
\begin{itemize} % Look at X-RAY: https://arxiv.org/pdf/2004.09039.pdf
    \item Exactly one instance of the search target exists in the environment.
    \item All objects are extruded polygons without stacking.
    \item Toppling does not occur.
    %\item All objects start and remain in stable poses
    \item The target object is not moved.
    %\item The camera is not moved 
    %\item Objects are not removed from the lateral access environment 
    %\item Only pushes are used to move objects, no grasps are used
\end{itemize}

We represent the states of the system $s(t)= \{Z_t, p_{t}, p'_{t-1}\}$, where $Z_t \in \mathbb{R}^{w\times h}$ is the current depth image,  ${p_{t}(x,y)}$ is the current 2D predicted target occupancy distribution, and ${p'_{t-1}(x,y)}$ is the 2D history minimum predicted target occupancy probability computed at the previous time step. Both ${p_{t}(x,y)}$ and ${p'_{t-1}(x,y)}$ are outputs from a perception pipeline we will introduce in 
\ref{sec:perception}.

The robot takes %has the action $a(t) \in \mathcal{A}$, where $\mathcal{A}$ is the set of all
actions $a(t) = (D,d,(x,z))$, where $D\in \{L,R\}$ is the pushing direction (left or right), and $d \in \mathbb{R}^+$ is the pushing distance. The coordinate $(x,z)$ is the pushing starting point in the camera frame.  

We define $P(x,t)$, a 1D occupancy distribution and the cost function $c(t)$, the negative cross entropy of the occupancy distribution as follows,
\begin{align}
    p'_t(x,y) &= \min{(p_t(x,y),p'_{t-1}(x,y))},\\
    P(x,t) &= \sum_{y=0}^{h-1} p'_t(x,y),\\
    c(t) &= -\sum_{i=0}^{w-1} P(x_i,t) \log(P(x_i,t)),
\end{align}
where $h,w$ is the depth image height and width.
%do we have to list it? can we just use something like (1)....
% some assumptions can be removed since they are already stated in the text.
%okay, let me work on reducing space.
% are there any other places that we can reduce space????
% Is there anyway to get the spaces between paragraphs/figures & text to be less?
% look at Fig 6 and the space below it, this happens in quite a few places, not sure if we can reduce that space, but may be a good try
% We're barely over 6 right?  This is easy.
% actually I just checked, we're right under 6
% wait actually we're over 6 (barely) wi acknowledgements  We can submit w/o ACKS.  Only the print version needs ACKs.  They ACKs should not effect the anon reviews.

%the problem of locating a known size target object with a stationary pose in a lateral-access environment with known dimensions by pushing occluding objects away. We assume all objects are extruded polygons without stacking and a fixed camera transformation. 

%notes: can't move target, push objects to avoid collision, single camera view, stable poses, avoid toppling, can't remove objects from shelf, left right pushes considered only, minimize number of steps taken, 

\section{Methods}
\label{sec:methods}

%\subsection{LAX-RAY}
We propose LAX-RAY, an automated system combining (1) a perception pipeline to predict the target object occupancy distribution with (2) a search policy utilizing this information to solve the problem efficiently.
%The perception pipeline starts with an RGBD image and computes a distribution of locations in the image behind which the target object may be occluded.  The search policy uses the output of the perception pipeline to perform an informed search of the shelf, computing the next pushing action the robot should take to attempt to reveal the target object.
The perception pipeline predicts an occupancy distribution for the (partially) occluded target object that spans across the visible objects in the scene given the depth image and target object segmentation mask. The search policy then computes the pushing action given a history of outputs from the perception module and the current observation at each step to efficiently uncover the target object.

\begin{figure}[t!]
\vspace{4pt}
    \centering
    \includegraphics[width=0.48\textwidth]{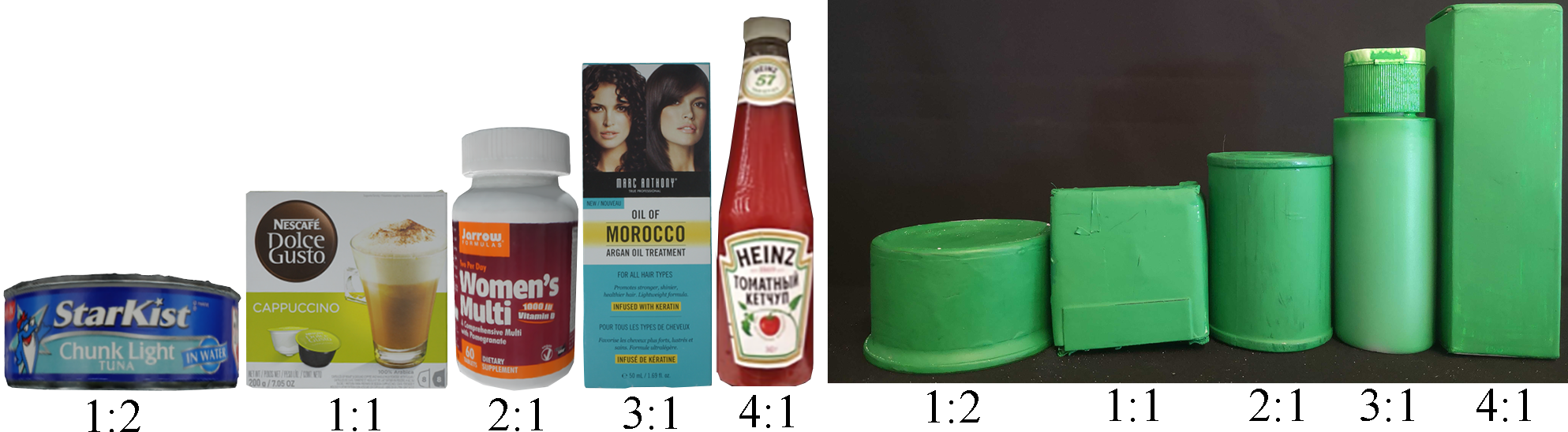}
    \caption{Target objects used in simulation testing (left) and in physical experiments (right).}
    \label{fig:targets}
    \vspace{-12pt}
\end{figure}

\begin{figure*}[t]
\includegraphics[width=\textwidth]{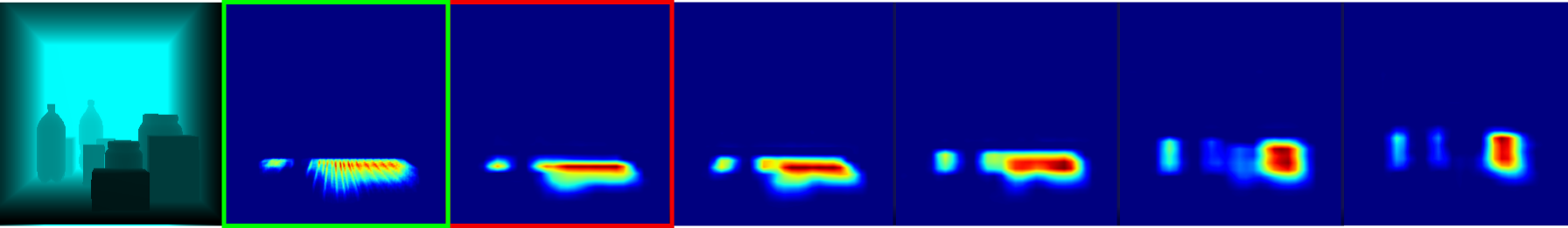}
 \caption{Simulation validation results from test set for target object with 1:2 aspect ratio in a fully occluded case, where the left (first) image is the depth observation in simulation, the second image in \textbf{green} box is the ground truth occupancy distribution and the third image in \textbf{red} box is the prediction output corresponding to the correct target aspect ratio. The fourth to the last images shows the network prediction given aspect ratio from 1:1 to 4:1 respectively. Significant difference between the predictions implied the critical influence of target aspect ratios.}
    \label{fig:xray_sim}
\end{figure*}

\begin{figure*}[t]
\vspace{4pt}
\includegraphics[width=\textwidth]{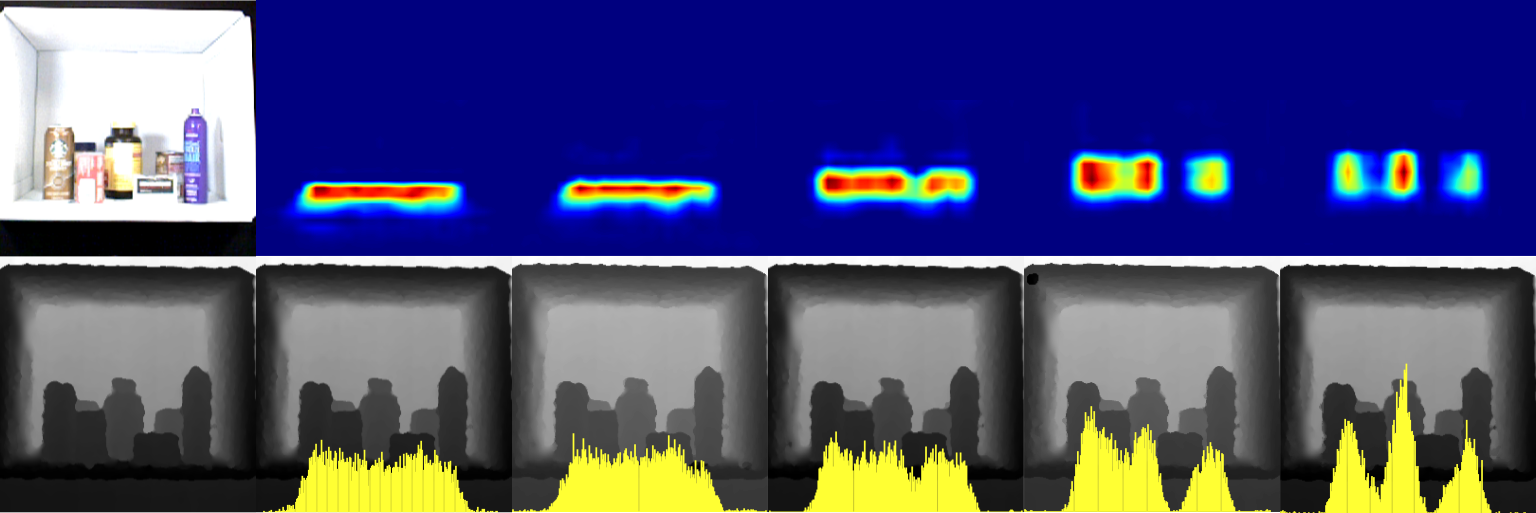}
 \caption{Validation of the pre-trained model on physical experiment environment for 5 different target object with aspect ratios from 1:2 to 4:1 in a fully occluded case. The left column shows the color observation of the shelf with random objects and the corresponding depth image taken by the PrimeSense camera. The top row shows 2D occupancy distribution from the pre-trained model and the second row shows the corresponding 1D occupancy distribution along X axis in camera frame overlaid with the depth observation. A significant difference for predictions of each aspect ratio is shown validating the accuracy of the pre-trained model for real applications. }
     \label{fig:xray_real}
     \vspace{-12pt}
\end{figure*}

\subsection{LAX-RAY Perception}
\label{sec:perception}
To predict an occupancy distribution for occluded and partially occluded objects from an depth image, we pre-train a neural network on synthetic data.
This model predicts the target object's location in the image plane.
%We get a projected predicted occupancy distribution for the target object's location on the X-Y plane of camera frame 
% (illustrated in Figure~\ref{fig:real}) 
%from observations of the current environment using a pre-trained model.

%We use simulation to generate a large amount of training data.
To generate large amounts of training data, we use a simulation (based on Trimesh~\cite{trimesh}) with the constraints introduced by the lateral access environment. 
%Including that the objects are in stable poses on the bottom plane without stacking. 
To generate data, the simulation first puts the target at a uniformly random position on a shelf. Occluding objects are then placed randomly on the shelf with no collisions. Due the aspect ratio's dominating effect on the occupancy distribution we used five cuboid target objects with varying aspect ratios (from 1:2 to 4:1 shown in 
Fig.~\ref{fig:targets}) to train for varying target objects.

%To explore a large number of possible positions of the target objects and generate the occupancy distribution, we produce a base map including the possible transformations of the target under the pose constraints, for the generation of the ground truth training data.
For each target object, we uniformly sampled transformations in the camera frame, with $14$ translations in the $x$ direction, $16$ translations in the $z$ direction, and $8$ rotations in the $x$-$z$ plane. This gives us 1792 transformations per target object, which empirically works well.
% More transformations don't improve the network performance significantly empirically.  are 

% Once the simulated shelf environment is populated, we render a 256$\times$256 depth image using iGibson~\cite{xia2020interactive}. The target modal mask at each transformation generated above is compared with the modal mask of the target in this scene. The amodal target masks of the target that produce the same modal mask (as seen in the scene) are added to the ground truth target occupancy distribution, which is then normalized. 

After populating the simulated shelf environment, we render a 256$\times$256 depth image and the corresponding target segmentation mask using iGibson~\cite{xia2020interactive}. We use the latter in conjunction with the target segmentation mask for each possible transformation above to generate the ground truth occupancy distribution over all possible transformations as in~\cite{danielczuk2020x}.
% We use the latter with the target segmentation mask for each possible transformation above ...
%The simulation data enables us to explore and learn possible positions of target objects in a lateral access environment using the depth observation. 
To mitigate the uncertainty of the camera position in physical experiments, we render observations from 5 random (uniformly sampled) camera positions for each populated simulation environment.
%Additionally, to mitigate the uncertainty of the camera position in the physical environment, for each scene, we take observations from 5 random different camera positions sampled from a uniform distribution around the center of the shelf.

% To conduct a search in the lateral environment and to optimize for efficiency, our approach takes in consideration the characteristics of depth observations produced from objects organized into random configurations on a simulated shelf in the X, Z plane.
% \textcolor{red}{Figures about coordinate frames to be added per KG Oct.24 feedback} 

% The size of our observational depth images is $256\times256$ pixels. We apply $14$ translations in the X direction and $16$ translations in the Y direction and $8$ rotations in the X-Z plane to the target object and compare the modal mask of the target at each transformation with the modal mask of the target in the original image. The amodal target masks corresponding to the transformations that have matching modal masks are added to the ground truth target occupancy distribution, which is normalized. We find empirically that considering more than the $1792$ total transformations per target object used here does not improve the network performance significantly and is computationally expensive.

We render over 30,000 images of training data for 5 target objects with occluding objects from the Google Scanned Objects dataset~\cite{scanned} in total. About 50\% of these images include a fully occluded target object. We split the data into training and validation sets with a 4:1 ratio. We also render 10000 images from a separate group of object models with different target objects with similar aspect ratios as the test set as shown in Fig. \ref{fig:targets}. The occluding objects in the test set have different categories with different shapes.
% The test target objects have similar aspect ratios and sizes but different shapes 

We train a fully-connected network (FCN) with a ResNet backbone with 33 million parameters on the rendered data set using stochastic gradient descent with a momentum 0.99 for 60000 iterations with batch size 32 and learning rate $10^{-5}$. This network takes input of the target object segmentation mask as well as the depth image of the current shelf and outputs the occupancy distribution density map of the target object. 
% \textcolor{red}{Most likely will not have space for learning curve/plots of error on test set unless they are put in the appendix}.

% {\color{red}TODO: We need a more high level para which will say what the challenges are of using x-ray and how we solve it here. We also want to say the main steps we're going to take - i.e. generate the occupancy distribution from image point of view, then generate a 3d or depth-aware representation of it, and etc}

\subsection{Search Policies}
%We model the lateral access mechanical search problem as a partially observable Markov decision process.

% {\color{red}[Is there a reason to have $(x,y)$ next to $z_t$?  Can we change notation as I did here for $z_t$?  Set notation will help shape and scope of the variables. -JI]}

 The search policy computes an action $a(t)=(D,d,(x,z))$ defined above to take at each step to find the target object. 
% To compute the action at time $t$, it consider the current depth image $z_t \in \mathbb{R}^{w\times h}$, the current predicted target occupancy distribution ${p_{t}(x,y)}$, and the history minimum predicted target occupancy probability ${p'_{t-1}(x,y)}$ computed at the previous time step, implying the state $s(t)= \{z_t, p_{t}, p'_{t-1}\}$. We consider each pushing action to be a tuple of start and end blade positions $a^-_t, a^+_t \in \mathbb{R}^3$. 
% {\color{red} [Is this accurate?  I thought it was a contact point ($\mathbb{R}^2$), direction (${L,R}$), and distance ($\mathbb{R}^+$).  Or more to the point, we need 1 depth. -JI]} 
A segmentation mask, generated based on significant depth discontinuities present in lateral access, is used to determine starting point candidates $x$. We restrict the pushing starting point $y$ in camera frame to be a fixed relatively low height to avoid toppling.
% To compute the start and end positions, segmentation masks are used of our objects which provide us with taking advantage of major discontinuities in depth value of the observation as well as the estimated collision free distance. Actions using these start and end positions are illustrated in Figure~\todo{missing fig}.

% We define an action $a(t)$, the current step 2D occupancy distribution $p_t$, history minimum 2D  occupancy distribution $p'_{t-1}$, 1D occupancy distribution $P(x,t)$ and cost $c(t)$ as follows,
% \begin{align}
%     a(t) &= \{a^-_t, a^+_t\},\\
%     p'_t(x,y) &=min{(p_t(x,y),p'_{t-1}(x,y))},\\
%     P(x,t) &= \sum_{y=0}^{h-1} p'_t(x,y),\\
%     c(t) &= -\sum_{i=0}^{w-1} P_t(x_i,t) log(P_t(x_i,t)).
% \end{align}
% where $h,w$ is the depth image height and width, $P(x,t)$ is a 1D  occupancy distribution along X axis in camera frame of the minimum between the current step 2D occupancy distribution $p_t$ and the history minimum 2D occupancy distribution $p'_{t-1}$, and $c(t)$ is the negative cross entropy of the occupancy distribution, as we want to minimize the predicted distribution support. $p'_{t-1}$ allows us to utilize the prior information. 
The blade insertion depth $z$ is directly proportional to the width of the object being pushed. The horizontal pushing distance $d$ is the distance the object can be pushed until collision calculated by a highly negative depth gradient from $Z_t$, which is unique for each pair of $(D,(x,z))$. Thus, for each object, our policy aims to explore which object to push, implied by $x$ and which direction $D$ to push.
%The horizontal pushing distance is determined by the maximum distance between the start position and an area of potential collisions, indicated by a highly negative depth gradient in the depth observation image.

Using the occupancy distribution from the perception pipeline, we define the following search policies to locate the target object efficiently:% Distribution Area Reduction Policy (DAR), Distribution Entropy Reduction Policy (DER) and Distribution Entropy Reduction over Multiple Time Steps (DER-MT) with a comparative uniform policy.

\begin{description}[align=left,leftmargin=0pt]
% \begin{itemize}
% \item[Uniform] is a policy that ablates DAR by removing the LAX-Ray prediction to show LAX-Ray's effectiveness.  
% The target occupancy distribution at time step $t$ is formulated using the indicator variable $\mathbbm{1}_x(t)$, that is 1 when an object is detected at X-axis value $x\in X$ (where $X$ is the set of X-axis values) and 0 otherwise: 
% $$P(x,t) = \min\left(\frac{\mathbbm{1}_x(t)}{\sum_{x\in X}\mathbbm{1}_x(t)}, P(x,t-1)\right)$$
% Let $P(x,-1)=1$ for all $x$. 
\item[Distribution Area Reduction (DAR)] ranks available actions using the current depth image and for every object computes the summed overlap of the object mask and the 1D minimum predicted occupancy distribution $P(x,t)$. The policy selects the action that reduces this sum the most. If the target object is partially revealed, the policy takes no action that would cover the target more.
% \item[Distribution Entropy Reduction (DER-1)] executes the action that minimizes the negative entropy $c(t+1)$ of the predicted distribution $\hat{P}(x,t+1)$ at the next time step, to minimize the predicted distribution support. If the target object is partially revealed, we restrict DER to not select actions that would cover the partially occluded target object more than it was in the previous step.

% hacky fix to no-line-break problem, split off some of the \item[] contents into a \textbf
\item[Distribution Entropy Reduction over n steps] \textbf{(DER-n)} computes the 1D predicted distribution $\hat{P}_{t+n}$ after taking $n$ actions and chooses the action that it predicts to produce the smallest entropy value $\hat{c}(t+n)$ at time step $t+n$ if optimal actions are taken given the current information, where $n\geq 1$. It predicts $\hat{P}_{t+n}$ by first predicting the depth image through translating the depth values of the object segmentation mask on the current depth image with the assumption of no other objects are behind. Then we get the occupancy distribution for this new depth image to get the predicted states. 

\item[Uniform] is created by substituting the predicted occupancy distribution from DAR with a uniform distribution for places with occluding objects. It does not benefit from the perception pipeline.
% \item[DER over Multiple Time steps (DER-MT(n))] %computes a tree of all actions across $n$ time steps, 
% is an extension of DER-1 that analyzes actions over multiple time steps. It computes the predicted distribution $\hat{P}_{t+n}$ after taking $n$ actions and chooses the action that it predicts to produce the smallest entropy value $\hat{c}(t+n)$ at time step $t+n$ if optimal actions are taken given the current information, where $n\geq 1$. 
% Additionally, we restrict actions that cover the target object more than in previous steps from being considered. {\color{red} [I think something is misworded here, how can we restrict actions that cover the target object if we don't know where the target object is? -JI]}
% \vspace{-3pt}
\end{description}
\begin{figure}[ht!]
\vspace{-4pt}
    \centering
    \includegraphics[width=\columnwidth]{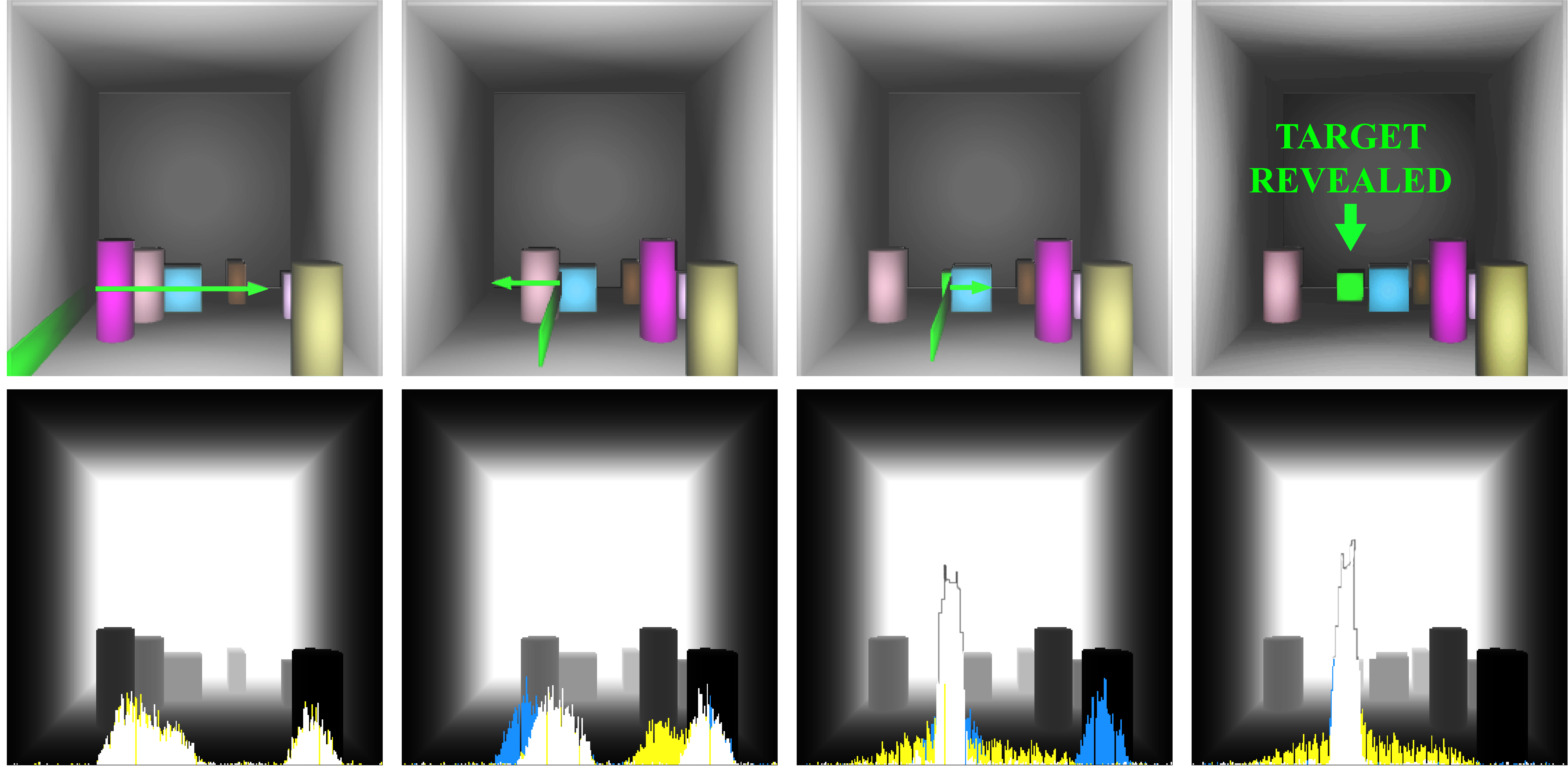}
    \caption{The DER-1 policy reveals a hidden object with 6 occluding objects. First row shows color images at each step with a green blade, and a green arrow denoting push direction and distance. The occupancy distribution at the bottom of each depth image (second row) includes: predicted distribution from previous time step (\textbf{blue}), predicted distribution at current time step (\textbf{yellow}), and minimum of the two (\textbf{white}). Target is revealed in last image, where the occupancy distribution has a very dense probability.}
    % \textbf{Left:} the green target cube before occluding objects are placed.
    % \textbf{Left-to-right:} 
    % DER reduces entropy by producing actions to narrow $P(t+1)$. }
    
    % \textbf{Right:} DER reveals the target object. 
    % \JI{Can we remove the left-most figure?  I think it may do more harm than good, and it would be easier to label each column with Left: Middle: and Right:.  Also, the spike in the distribution on the right is washed out.  Is it possible to outline the distribution so it shows up?  Finally, can you put the text "Target Revealed" above the green box with an arrow pointing to the target in the last figure.}
    \label{fig:DER_success}
    \vspace{-12pt}
\end{figure}

% \rule{0pt}{8pt}\\
\begin{figure*}
\vspace{4pt}
     \begin{subfigure}[t]{0.16\textwidth}
    \includegraphics[width=\textwidth]{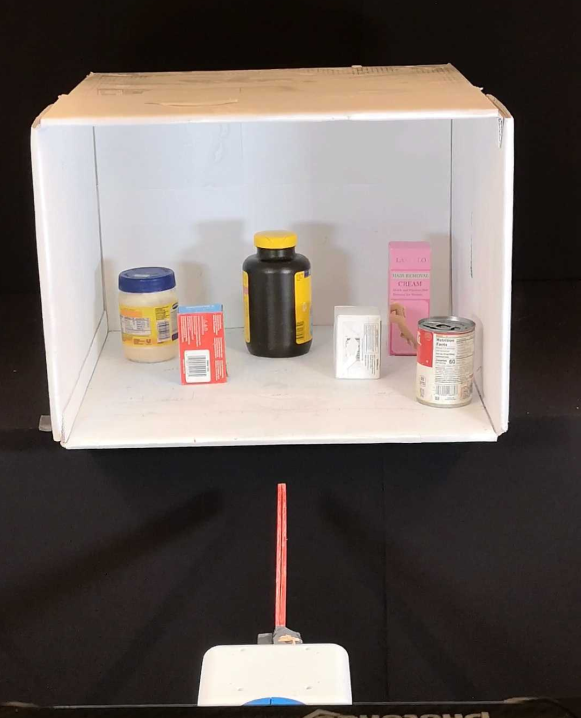}
    \caption{Initial position before push.}
    \label{fig:push1}
    \hfill
    \end{subfigure}
         \begin{subfigure}[t]{0.16\textwidth}
    \includegraphics[width=\textwidth]{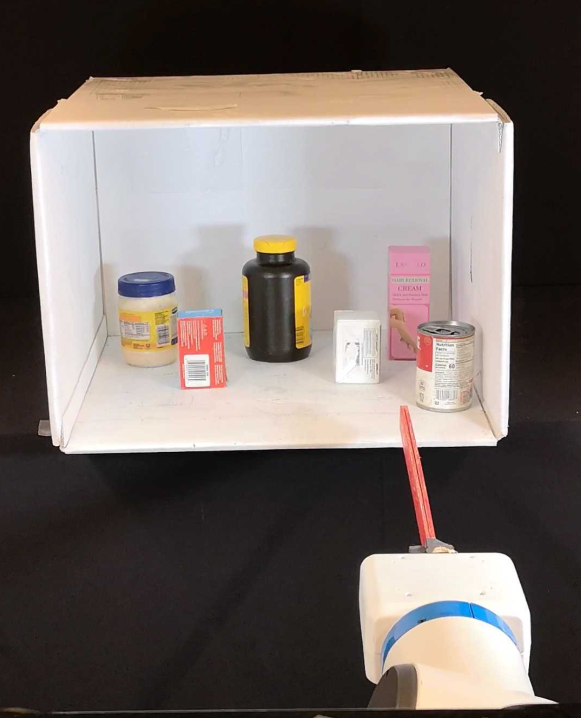}
    \caption{Move near the push starting position.}
    \label{fig:push2}
    \hfill
    \end{subfigure}
         \begin{subfigure}[t]{0.16\textwidth}
    \includegraphics[width=\textwidth]{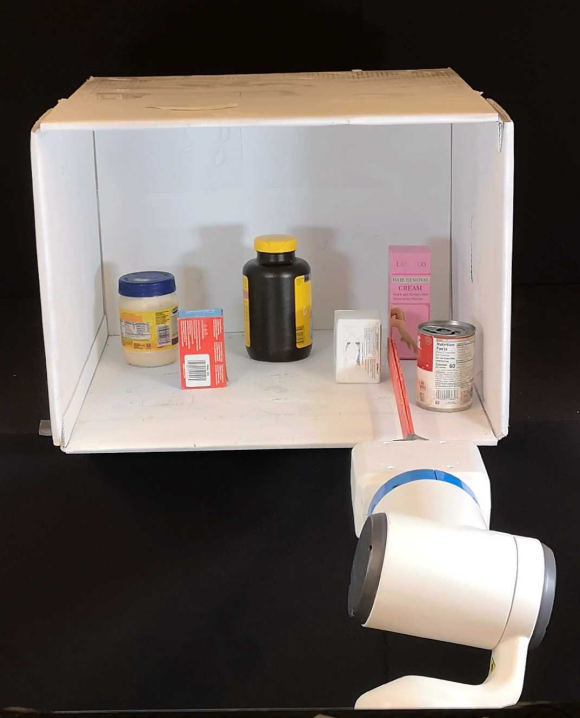}
    \caption{Insert to the push starting position.}
    \label{fig:push3}
    \hfill
    \end{subfigure}
         \begin{subfigure}[t]{0.16\textwidth}
    \includegraphics[width=\textwidth]{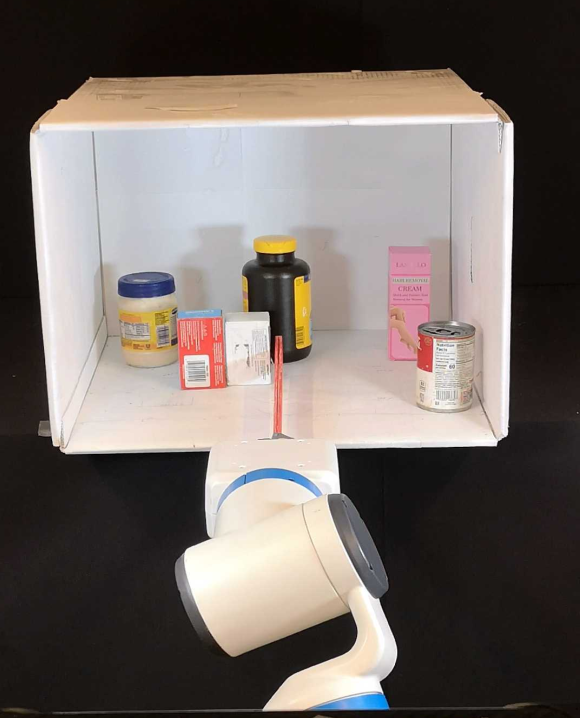}
    \caption{Execute pushing.}
    \label{fig:push4}
    \hfill
    \end{subfigure}
         \begin{subfigure}[t]{0.16\textwidth}
    \includegraphics[width=\textwidth]{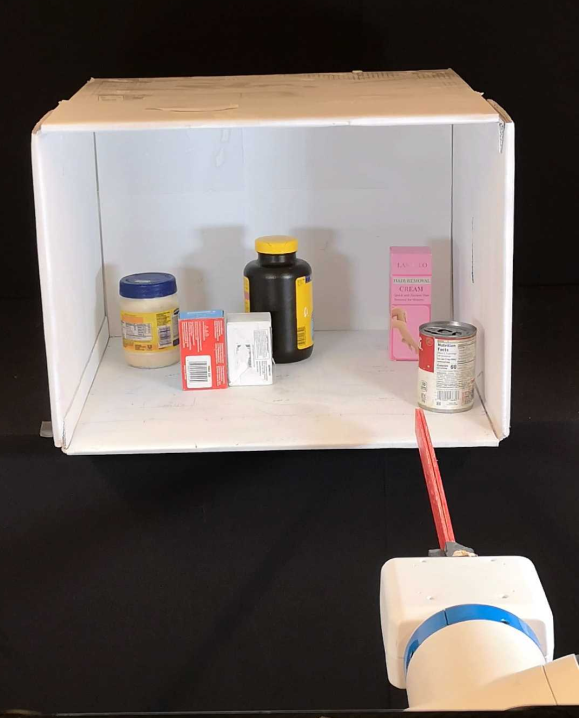}
    \caption{Goes back to the position at \ref{fig:push2}}
    \label{fig:push5}
    \hfill
    \end{subfigure}
         \begin{subfigure}[t]{0.16\textwidth}
    \includegraphics[width=\textwidth]{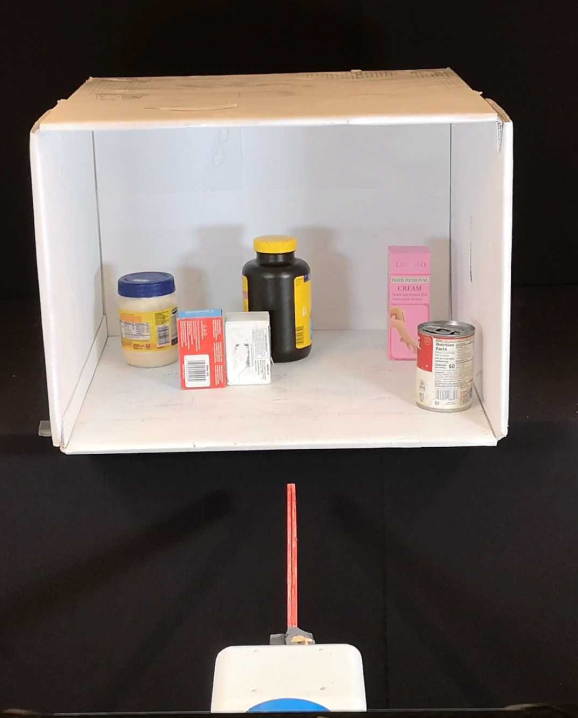}
    \caption{Goes back to the initial position after push.}
    \label{fig:push6}
    \hfill
    \end{subfigure}
    \vspace{-12pt}
\caption{5 pushing steps for each pushing action.}
    \label{fig:push}
\end{figure*}

\begin{figure*}[t]
    \centering
    \includegraphics[width=\textwidth]{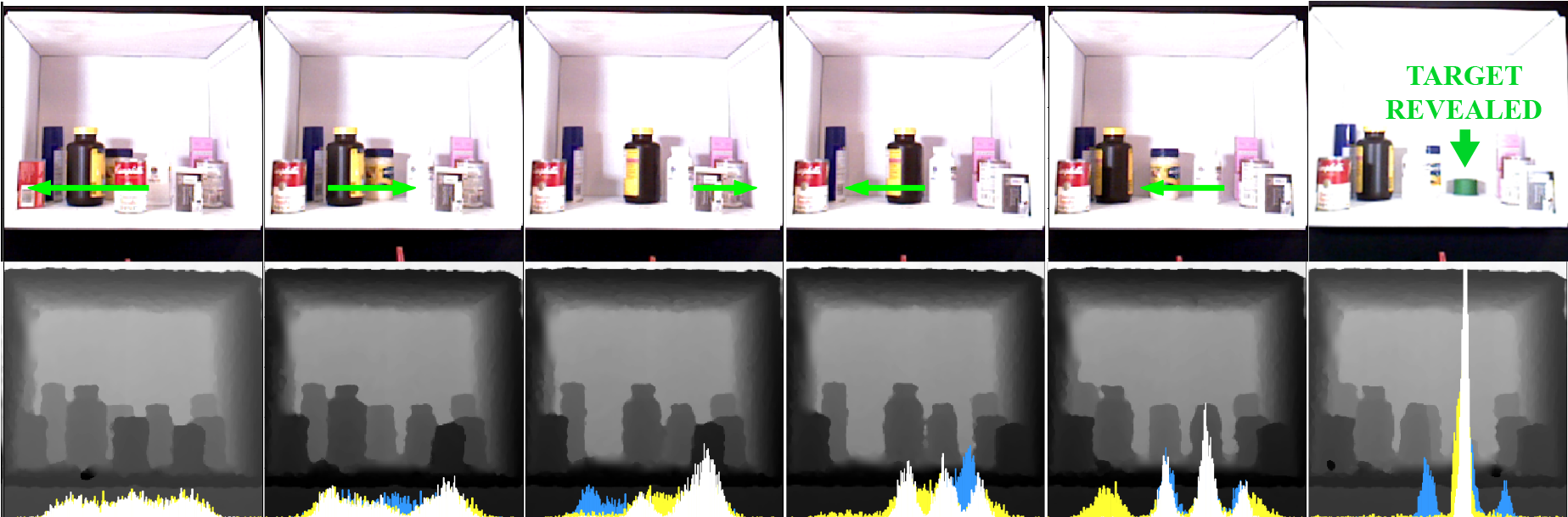}
    \caption{The DER-1 policy reveals the target object among 9 occluding objects in a physical experiment. The plotted occupancy distributions at the bottom of each depth image including three parts: the predicted distribution from the previous time step (blue), the predicted distribution at the current time step (yellow), and minimum of the two distributions (white).  \textbf{Left:} the target position and depth image. Each subsequent column shows the action output over time.  \textbf{Top:} RGB image with green push action arrow denoting the pushing direction and distance.  \textbf{Bottom:} depth image with occupancy distributions.}
    \label{fig:phy_action}
    \vspace{-12pt}
\end{figure*}
\section{First Order Shelf Simulator (FOSS)}
\label{foss}
To validate the proposed policies, we propose a First Order Shelf Simulator (FOSS), a Python-based simulator that uses Trimesh~\cite{trimesh} and Pyrender to model object-object interactions within a parametrizable shelf environment. 
% The shelf is a  cube with edges of size 0.6 meters, with the lateral face of the cube missing to mimic a shelf. 

FOSS represents objects as extruded polygons (cuboids and cylinder approximations) with pose constraints matching the problem statement. Cuboids uniformly range from 0.02 to 0.10~m for the square base and are 0.1~m tall. Cylinders uniformly range from 0.02 to 0.05~m for the base radius and are between 0.1 and 0.2~m tall, also sampled uniformly. A green target object cube with sides of size 0.07~m and aspect ratio of 1:1 is used. Action $a(t)$ given by the policy is executed without friction.   
% Actions are defined by an object to push, a 2D push direction vector, and a push distance without friction
FOSS checks collisions between targets and shelf wall. If an object collides with another object during the push, both objects move together for the remainder of the push in the push direction. When an object collides with the shelf wall, the push action stops.

Although FOSS allows for pushes in 2D, we focus on 1D lateral pushes in this paper. For simulation runs we place objects randomly on the shelf while checking for collisions to ensure object separation. Furthermore, to guarantee possible initial pushes, objects are initialized with at least the blade thickness away from the wall. Ground truth segmentation masks are used in FOSS.
% We generated random scenes and for scene, a
% The simulation is completed if the target object is at least 90\% exposed. 
% {\color{red}TODO (Vishal + Marcus): More detail w.r.t. code, ex. Github, Python, etc.}
\section{Experiments} \label{sec:experiments}
% {\color{red}TODO: Need more details everywhere.}
We tested the perception system and the proposed policies both in simulation and physical environment.
\subsection{Perception Experiments in Simulation}
%\begin{table} [!tbp]
%\centering
%\scalebox{.92}{
%\begin{tabular}{|c|c|c|c|} \hline
%Validation Loss & Validation IoU & Testing Loss &Testing IoU \\ [.2ex] 
%\hline
%80.5 & 0.79&193.3&0.53\\
%[0ex]
%\bottomrule
%\end{tabular}}
%\caption{Bench marking results}
%\label{tab:fcn}
%\end{table}

In the simulation environment, images from both validation and test sets are used to benchmark the model, yielding a validation IoU of 0.79 and a test IoU of 0.53. Example images from the test set are shown in Fig. \ref{fig:xray_sim} in the case where the target object is fully occluded. The predicted distributions vary significantly across target object aspect ratios, indicating that the distributions are particularly sensitive to the aspect ratio.
\subsection{Perception Experiments in Physical}
In physical experiments, we use green target objects with similar sizes and aspect ratios to those used in simulation, including cuboids and cylinders, as shown in Fig. \ref{fig:targets}. We put random objects on a white shelf and use a PrimeSense RGBD camera embedded on a Fetch robot to get the color and depth images. We use a color detection algorithm on the color image by setting thresholds for RGB values to detect the target object, which is painted green.
% as shown in Fig. \ref{fig:target_phys}. 
Fig. \ref{fig:xray_real} shows the prediction results of the pre-trained model on a shelf with randomly arranged objects in a fully occluded case for all five target objects.

% \begin{figure*}[t]
%     \centering
%     \includegraphics[width=1.\textwidth]{includes/figures/dists3.png}
%     \caption{LAX-RAY target occupancy distribution predictions for simulated and real observations. Row (A) depicts LAX-RAY outputs given the leftmost image and a target aspect ratio between 1:2 and 4:1 as input. Rows (B) and (C) show the same for the real observation (the leftmost image in B) as both heatmaps (row (B)) and PDFs (row (C)).}
%     \label{fig:real}
% \end{figure*}
\begin{table}[ht!]
	\vspace{4pt}
	\centering
	\caption{Simulation results for 5 policies over the same 200 simulation setups. For different number of occluding objects, the first row is the \textbf{Success Rate} followed by the \textbf{Average} of Steps and \textbf{Standard Deviation} of steps. As DER-n assume no unseen objects are behind the pushed object, DER-n policies have higher prediction errors as n increases.  Addressing this is an important topic for future research. }

	%Raven, look good?
	%due to the assumption of no unseen objects behind the object to be pushed, DER policies accumulate prediction errors with more prediction steps.  % looks good
%The prediction of the next step assumes no unseen objects are behind the pushed object, which can deviate from what actually happens. Thus, there is a tradeoff between the prediction errors and lookahead. Compared to DER-MT2, DER performs worse because not enough future information is revealed while DER-MT3 performs worse because prediction error accumulates. When there are fewer objects, prediction errors dominate this tradeoff, which could explain the better performance of DAR compared with DER-MT2.
	\vspace{0.5mm}
\begin{adjustbox}{width=\columnwidth,center}
    \centering
\begin{tabular}{c|c|c|c|c|c|c}
    \hline
    No.&Results&Uniform&DAR&DER-1&DER-2&DER-3\\
    \hline
    \hline
  \multirow{2}*{2}&Succ.&97$\%$&\textbf{98$\%$}&\textbf{98$\%$}&97$\%$&94$\%$\\
&Step Avg&1.34&1.36&1.36&1.80&1.99\\ 
&Step Std&0.56&0.77&0.67&0.93&1.19\\
\hline
  \multirow{2}*{4}&Succ.&88$\%$&\textbf{96$\%$}&92$\%$&92$\%$&89$\%$\\
&Step Avg&2.27&2.33&2.37&2.57&3.21\\ 
&Step Std&1.48&1.78&1.79&1.93&2.02\\
\hline
  \multirow{2}*{6}&Succ.&71$\%$&87$\%$&79$\%$&\textbf{89$\%$}&85$\%$\\
&Step Avg&2.99&2.79&3.04&3.32&3.82\\ 
&Step Std&2.21&2.06&2.30&2.14&2.44\\
\hline
  \multirow{2}*{8}&Succ.&46$\%$&66$\%$&63$\%$&\textbf{71$\%$}&66$\%$\\
&Step Avg&3.61&3.40&3.13&3.94&4.08\\ 
&Step Std&2.75&2.47&2.25&2.65&2.54\\
\hline
  \multirow{2}*{Avg}&Succ.&75.5$\%$&86.8$\%$&83.0$\%$&\textbf{87.3$\%$}&83.5$\%$\\
&Step Avg&2.56&\textbf{2.47}&2.48&2.91&3.28\\
\hline
\end{tabular}
\end{adjustbox}
	\label{tab:sim_result}
	\vspace{-12pt}
\end{table}
% \begin{table*}[h!]
% 	\centering
% 	\caption{Simulation results for 5 policies over the same 200 simulation setups. The columns are organized such that for each number of objects, we have the \textbf{Percent Success} of when we observe 90\% of the target and the \textbf{25th Quartile}, \textbf{50th Quartile}, and \textbf{75th Quartile} of number of actions to succeed. }
% 	\vspace{0.5mm}
% 	\begin{tabu} to \textwidth {X[3l]X[1.5c]X[c]X[c]X[c]X[1.5c]X[c]X[c]X[c]X[1.5c]X[c]X[c]X[c]X[1.5c]X[c]X[c]X[c]} \toprule
% 	  & \multicolumn{4}{|c}{\textbf{2 Objects}} & \multicolumn{4}{|c}{\textbf{4 Objects}} & \multicolumn{4}{|c}{\textbf{6 Objects}} & \multicolumn{4}{|c}{\textbf{8 Objects}} \\
% 		\textbf{Policy} & \multicolumn{1}{|c}{Succ.} & \multicolumn{3}{c}{Action Quartiles} & \multicolumn{1}{|c}{Succ.} & \multicolumn{3}{c}{Action Quartiles} & \multicolumn{1}{|c}{Succ.} & \multicolumn{3}{c}{Action Quartiles} & \multicolumn{1}{|c}{Succ.} & \multicolumn{3}{c}{Action Quartiles} \\ \midrule
% 		Uniform & 0.97 & 1 & 1 & 2 & 0.88 & 1 & 2 & 3 & 0.71 & 1 & 2 & 4 & 0.46 & 1 & 3 & 5 \\
% 		DAR & \textbf{0.98} & 1 & 1 & 2 & \textbf{0.96} & 1 & 2 & 3 & 0.87 &1 & 2 &4 & 0.66 & 1 &3 &5\\
% 		DER & \textbf{0.98} & 1 & 1 & 2 & 0.92 & 1 & 2 & 3 & 0.79 & 1 & 2 & 4 & 0.63 & 1 & 2 & 4\\
% 		DER MT2 & 0.97 & 1 & 2 & 2 & 0.92 & 1 & 2 & 3 & \textbf{0.89} & 2 & 3 & 4 & \textbf{0.71} & 2 & 3 & 6 \\
% 		DER MT3 & 0.94 & 1 & 2 & 2 & 0.89 & 1 & 3 & 5 & 0.85 & 2 & 4 & 5 & 0.66 & 2 & 4 & 5 \\\bottomrule
% 	\end{tabu}
% 	\label{tab:perceptionbenchmark}
% 	\vspace{-6pt}
% \end{table*}
\begin{table}[t!]
\vspace{4pt}
% \rule{0pt}{2pt}\\
    \caption{Physical experiment results. The first column shows the number of occluding objects.  The remaining columns show \textbf{task success (top)} with Y denoting success and N denoting fail and the number of \textbf{actions to succeed (bottom number)}.  The last row shows the average of the 5 experiments.}
\begin{adjustbox}{width=\columnwidth,center}
    \centering
\begin{tabular}{c|c|c|c|c|c|c}
    \hline
    No.&Results&Uniform&DAR&DER-1&DER-2&DER-3\\
    \hline
    \hline
  \multirow{2}*{2}&Success&Y&Y&Y&Y&Y\\
&Steps&3&3&3&1&1\\ 
\hline
  \multirow{2}*{4}&Success&Y&Y&Y&Y&Y\\
&Steps&1&1&2&1&1\\ 
\hline
  \multirow{2}*{6}&Success&Y&Y&Y&Y&Y\\
&Steps&3&4&4&6&6\\ 
\hline
  \multirow{2}*{7}&Success&N&N&Y&Y&Y\\
&Steps&4&7&3&4&3\\ 
\hline
  \multirow{2}*{9}&Success&N&Y&Y&Y&Y\\
&Steps&9&5&5&3&5\\ 
\hline
  \multirow{2}*{Avg}&Success&60$\%$&80$\%$&\textbf{100}$\%$&\textbf{100}$\%$&\textbf{100}$\%$\\
&Steps&4&4&3.4&\textbf{3}&3.2\\
\hline
\end{tabular}
% \newcommand{\Y}{{\footnotesize success}}
% \newcommand{\N}{{\footnotesize \emph{failed}}}
% \begin{tabular}{c|c|c|c|c|c}
%     \hline
%     No.&Uniform&DAR&DER&DER-MT2&DER-MT3\\
%     \hline
%     \hline
%   \multirow{2}*{2}&\Y&\Y&\Y&\Y&\Y\\
% &3&3&3&1&1\\ 
% \hline
%   \multirow{2}*{4}&\Y&\Y&\Y&\Y&\Y\\
% &1&1&2&1&1\\ 
% \hline
%   \multirow{2}*{6}&\Y&\Y&\Y&\Y&\Y\\
% &3&4&4&6&6\\ 
% \hline
%   \multirow{2}*{7}&\N&\N&\Y&\Y&\Y\\
% &4&7&3&4&3\\ 
% \hline
%   \multirow{2}*{9}&\N&\Y&\Y&\Y&\Y\\
% &9&5&5&3&5\\ 
% \hline
%   \multirow{2}*{Mean}&60$\%$&80$\%$&100$\%$&100$\%$&100$\%$\\
% &4&4&3.4&3&3.2\\
% \hline
% \end{tabular}
\label{tab:phy_result}
\end{adjustbox}
\vspace{-12pt}
\end{table}

\subsection{Mechanical Search Policy Experiments in Simulation}
%We tested the three policies and the uniform policy in both simulation and physical environments.

We use FOSS to rapidly test our policies in many scenarios. For this, we use a randomly generated assortment of cylinders and boxes, such that their positions and sizes are random within the shelf and the range specified in Section \ref{foss}. All objects satisfy the pose constraints and are on the floor of the shelf in stable poses.

%Additionally these objects are taller than they are wide to mimic objects typically found on shelves. We have a blade that we control to push the objects. For this we simulate simple lateral pushes with simple, frictionless object-object dynamics where we simulate both object-object and object-blade collisions. Pushes are given by the policy given the color image, depth image, segmentation, and history probability function from the simluated environment. The pushing action is defined as a start and end point in the camera frame where $x$ points to the right, $y$ points down, and $z$ points forwards, with $y$ and $z$ staying constant between the start and end point. To simulate pushes, we first test if a blade can be entered from the outside of the shelf at starting position $S_s = (x_s,y_s,z_s)$ through a linear motion, similar how it would happen in real. We then teleport the blade to this location and slide it linearly to the ending position $S_e=(x_e,y_e,z_e)$. We then teleport the blade out of the shelf, and go on to the next step. 

We generated 200 random scenes with 2, 4, 6, and 8 occluding objects respectively, giving us 800 scenes in total. We tested 3 prediction steps of DER-n for $\text{n}\in\{1,2,3\}$ together with Uniform and DAR on each scene. A green cube with a 1:1 aspect ratio is used for the target. A rollout is considered successful if at least 90$\%$ of the target object is revealed within 10 actions. The policies' performance in simulation is summarized in Table~\ref{tab:sim_result} with a successful DER-1 policy rollout in Fig.~\ref{fig:DER_success}. 

Table~\ref{tab:sim_result} suggests DAR and the DER-n policies perform better than the Uniform policy, especially as the number of objects increases. This shows the contribution of the occupancy distribution. All policies' performances drop when the number of occluding objects increases since more steps are needed to reveal the target. DAR performs best when there are fewer than 6 occluding objects and DER-2 perform best in 6 or more objects scenarios. The prediction of the next step assumes no unseen objects are behind the pushed object, which can deviate from what actually happens. Thus, there is a tradeoff between the prediction errors and lookahead. Compared to DER-2, DER-1 performs worse because not enough future information is revealed while DER-3 performs worse because prediction error accumulates. When there are fewer objects, prediction errors dominate this tradeoff, which could explain the better performance of DAR compared with DER-2. 

Most failure cases for DAR and DER-1 occur due to not looking ahead. Without lookahead, DAR and DER-1 struggle in cases where multiple layers of objects block the target; they often will avoid pushing the front object since it does not immediately change the predicted distribution. DER-2 and DER-3 mitigate this issue by considering the total reward of pushing both layers of objects. 

\subsection{Mechanical Search Policy Experiments in Physical}
To test the policies, we set up a shelf environment with random objects and execute the insertion and pushing actions using a Fetch robot with a blade attached to the gripper. An embedded PrimeSense camera is used for taking RGBD observations. Fig.~\ref{fig:phy_setup} shows the setup.
%The pushing action given by the policy is defined in the robot base frame where the $x$ axis is pointing forward, the $y$ axis is pointing to the left and $z$ axis is pointing upward. We get the blade pushing starting position $S_s = (x_s,y_s,z_s)$ and ending position $S_e=(x_e,y_e,z_e)$ from the policy given the color image, depth image, segmentation and the history probability prediction. At the fist step, we go from a fixed initial position $S_I = (x_I,y_I,z_I)$ to the pre-insertion position $S_p=(x_p,y_p,z_p)$ where $x_p=x_I, y_p=y_s, z_p=z_s$. At the second step, we execute the insertion action from the pre-insertion position to $S_s$. At the next step, we execute the pushing action from $S_s$ to $S_e$. After the pushing action, we go back from $S_e$ to $S_p$ and then go back to the initial position $S_I$. By executing those five steps, we avoid the collision between the blade and other objects. \textcolor{red}{Todo: Put failure/success examples of D}
% \begin{figure*}[tbh!]
%   \centering
%   \includegraphics[width=\textwidth]{includes/figures/pushing.png}
%   \caption{\textbf{Executing a physical push action}. Left-to-right: the robot in (a) its initial configuration, (b) translates the blade right, (c) inserts to the push contact point, (d) pushes the object left, (e) moves the blade out, and (f) resets to the initial configuration. \JI{It might be more clear if we had these as separate images that we could put subcaptions on}}
%     \label{fig:push}
% \end{figure*}
At each time step, we execute the pushing action returned by the policy in 5 steps starting and ending at the same fixed position as shown in Fig. \ref{fig:push}. By dividing the pushing action into those 5 steps, we avoid the occlusion of the robot arm to the camera and any potential collisions.

We tested 5 different layouts for each policy with 2 occluding objects, 4 occluding objects, 6 occluding objects, 7 occluding objects and 9 occluding objects respectively. A target object with aspect ratio of 1:2 from Fig. \ref{fig:targets} is used and is fully occluded in all layouts.

The performance of each policy is summarized in Table~\ref{tab:phy_result}. The results suggest consistency with the simulation results in which the uniform policy has the worst performance with the lowest success rate, and DER-2 shows the best performance with lowest average number of steps. 
% Though in the physical experiments, the policy failed since it would push the objects against the wall due to perception noise, we can still see more steps are taken by those failed polices compared with the successful policies. 
Fig.~\ref{fig:phy_action} shows an action series for experiment 5 with DER-1. 

% \begin{table}[t]
%  \caption{Physical experiment results for different policies. }
% \centering

% \begin{tabular}{c|c|c|c}
% \hline
% Experiment &Policy & Results & Number of Steps\\
% \hline
% \hline
% \multirow{5}*{1}&Baseline&Success&3\\
% &DAR&Success&3\\ 
% &DER&Success&3\\
% &DER-MT2&Success&1\\
% &DER-MT3&Success&1\\
% \hline
% \multirow{5}*{2}&Baseline&Success&1\\
% &DAR&Success&1\\ 
% &DER&Success&2\\
% &DER-MT2&Success&1\\
% &DER-MT3&Success&1\\
% \hline
% \multirow{5}*{3}&Baseline&Success&3\\
% &DAR&Success&4\\ 
% &DER-MT1&Success&4\\
% &DER-MT2&Success&6\\
% &DER-MT3&Success&6\\
% \hline
% \multirow{5}*{4}&Baseline&Fail&4\\
% &DAR&Fail&7\\ 
% &DER-MT1&Success&3\\
% &DER-MT2&Success&4\\
% &DER-MT3&Success&3\\
% \hline
% \multirow{5}*{5}&Baseline&Fail&9\\
% &DAR&Success&5\\ 
% &DER-MT1&Success&5\\
% &DER-MT2&Success&3\\
% &DER-MT3&Success&5\\
% \hline
% \end{tabular}
%  \label{tab:phy_result}
% \end{table}

% \begin{table}[]
% \centering
% \caption{Average Physical Experiment Results}
% \begin{tabular}{c|c|c}
% \hline
% Policy&Success Rate&Mean Number of Steps \\
% \hline
% Baseline&60\%&4\\
% % \hline
% DAR&80\%&4\\ 
% % \hline
% DER-MT1&100\%&3.4\\
% % \hline
% DER-MT2&100\%&3\\
% % \hline
% DER-MT3&100\%&3.2\\
% \hline
% \end{tabular}
% \label{tab:phy_result_ave}
% \end{table}

\section{Conclusion and Future Work} \label{sec:discussion}
% We acknowledge the limitations of our approach to mechanical search in lateral access environments. We make strong assumptions about objects and their configurations and do not consider more complex action sequences, such as those that involve backwards pushes. However, we believe that the methods proposed in this paper illustrate both the richness of the lateral access search problem as well as its feasibility.
This paper addresses the "lateral access" (shelf) variant of mechanical search and proposes three novel policies utilizing the additional constraints and the predicted target object occupancy distribution to optimize the number of pushing actions needed to reveal a target object on a shelf. 

In future work, we will investigate more sophisticated depth models and the use of pushes parallel to the camera (depth) axis to create space for lateral pushes, and pull actions using pneumatically-activated suction cups to lift and remove occluding objects from the shelf.  

\section*{Acknowledgements}
\footnotesize
This research was performed at the AUTOLAB at UC Berkeley in
affiliation with the Berkeley AI Research (BAIR) Lab. This research was supported in part by: NSF National Robotics Initiative Award 1734633 and by a Focused Research Award from Google Cloud. The authors were supported in part by donations from Google and Toyota Research Institute, the National Science Foundation Graduate Research Fellowship Program under Grant No. 1752814, and by equipment grants from PhotoNeo and NVidia. We thank our colleagues who provided helpful feedback and suggestions.

% \clearpage
\renewcommand*{\bibfont}{\footnotesize}
\printbibliography
\end{document}